\def\year{2018}\relax
\documentclass[letterpaper]{article} 
\usepackage{aaai18}  
\usepackage{amsfonts}
\usepackage{array}
\usepackage{booktabs}
\usepackage{times}  
\usepackage{helvet}  
\usepackage{courier}  
\usepackage{url}  
\usepackage{graphicx}  
\usepackage{amsmath}
\usepackage{tablefootnote}
\usepackage[linesnumbered,ruled,vlined]{algorithm2e}
\usepackage[utf8]{inputenc}

\usepackage{color}
\definecolor{midblue}{rgb}{0.2,0.2,0.7}
\definecolor{lightblue}{rgb}{0.15,0.3,0.8}

\newtheorem{definition}{Definition}

\begin{document}
%
\title{
On Cryptographic Attacks Using Backdoors for SAT
\thanks{A.\,Semenov, O.\,Zaikin, I.\,Otpuschennikov and S.\,Kochemazov are supported by Russian Science Foundation (project~16-11-10046).
I.\,Otpuschennikov is partially supported by the Council for Grants of the President of Russia (grant~SP-4751.2016.5).
A.\,Ignatiev is supported by FCT funding of post-doctoral grant SFRH/BPD/120315/2016 and LASIGE Research Unit, ref. UID/CEC/00408/2013.}
}
\author{Alexander Semenov$^1$, Oleg Zaikin$^1$, Ilya  Otpuschennikov$^1$, Stepan Kochemazov$^1$, Alexey Ignatiev$^{2,1}$\\
$^1$ Matrosov Institute for System Dynamics and Control Theory SB RAS, Irkutsk, Russia\\
$^2$ LASIGE, Faculdade de Ciências, Universidade de Lisboa, Portugal\\
}

\maketitle

\begin{abstract}
  Propositional satisfiability (SAT) is at the nucleus of state-of-the-art
  approaches to a variety of computationally hard problems, one of which is
  cryptanalysis.
  Moreover, a number of practical applications of SAT can only be tackled
  efficiently by identifying and exploiting a subset of formula's variables
  called backdoor set (or simply backdoors).
  This paper proposes a new class of backdoor sets for SAT used in the context
  of cryptographic attacks, namely guess-and-determine attacks.
  The idea is to identify the best set of backdoor variables subject to a
  statistically estimated hardness of the guess-and-determine attack using a
  SAT solver.
  Experimental results on weakened variants of the renowned encryption
  algorithms exhibit advantage of the proposed approach compared to the state
  of the art in terms of the estimated hardness of the resulting
  guess-and-determine attacks.
  %
  %
\end{abstract}

%
\section{Introduction}\label{sec:intro}

During the last two decades, algorithms for solving the Boolean satisfiability
problem (SAT) has become one of the key components of the state-of-the-art
approaches to computational problems from a multitude of practical domains
including artificial intelligence, software and hardware verification,
computation biology, among many others.
Nevertheless, there always remain exceptionally hard SAT instances that require
additional effort to be solved efficiently.
A common way to deal with a hard CNF formula is to attempt to determine
so-called \emph{backdoors} of the formula.
Here, a \emph{backdoor} or a \emph{backdoor set} is a subset of variables of
the formula that enables decomposing the original problem into a family of
``weakened'' subproblems, which are relatively easy to
solve~\cite{williams-ijcai03}.
Typically, solving the subproblems requires a polynomial-time algorithm.
A significant upside of backdoor sets is that they often make it possible to
estimate the time required to solve the problem.

An important practical area that gives birth to exceptionally hard
computational problems is cryptography.
Moreover, a number of attempts were made to tackle various cryptanalysis
problems using constraint satisfaction algorithms, e.g.\
SAT~\cite{mironov-sat06,de-sat07,erkok-plpv09,soos-sat09,tomb-sat15} and CSP~\cite{gerault-ijcai17}.

Most of the cryptanalysis problems can be formulated in the following way:
given a known encryption algorithm and its known output, one needs to find the
corresponding input of the algorithm.
There are a number of \emph{cryptographic attacks} aiming at solving such
problems, an important class of which comprises \emph{guess-and-determine}
attacks~\cite{bard-ac09}.
One may immediately notice an explicit relation between backdoor sets for SAT
and guess-and-determine attacks on cryptographic functions.
Indeed, they both follow a simple strategy: first, ``guess'' values of the
variables from some (backdoor) set, and then ``determine'' using a relatively
efficient algorithm whether or not the guess was correct.
Usually, guess-and-determine attacks are constructed as a result of a thorough
analysis of the features of the considered cipher and whether or not they can
be exploited by the intrinsic characteristics of the algorithm used to solve
the weakened cryptanalysis problems.

The present paper investigates application of the state-of-the-art SAT
technologies to automatically constructing guess-and-determine attacks on
relevant encryption algorithms.
Unfortunately, in most of the cases backdoors introduced in previous
works 
cannot be applied efficiently for this purpose.
The reason is that for every output of a cryptographic function one essentially
has to construct a separate backdoor set, i.e.\ it cannot be applied to a
family of problems.
Also, in some cases restrictions on the class of underlying algorithms (e.g.\
Strong Backdoors~\cite{williams-ijcai03} assume the use polynomial-time
procedures) are too severe.

The paper makes the following contributions.
First, the paper proposes a new type of backdoor sets for SAT called
\emph{Inverse Backdoor Sets} (IBS).
It extends the previously introduced notion of Strong Backdoors
\cite{williams-ijcai03} towards allowing one to use algorithms for solving
problems from NP as well as incomplete algorithms when solving
simplified/weakened subproblems.
Second, the paper introduces a special \emph{resistance function} used to
estimate a running time of the guess-and-determine attack corresponding to the
specific Inverse Backdoor Set.
Computation of the resistance function is done using the Monte-Carlo
method~\cite{metropolis-jasa49}.
Third, borrowing the ideas from \emph{blackbox optimization}, a general method
is proposed for identifying an IBS with a good runtime estimation of the
corresponding attack by minimizing the resistance function over the set of all
possible IBSes.
The fourth contribution of the paper is that the approach is applied to
constructing guess-and-determine attacks on several relevant cryptographic
systems, such as the Trivium keystream generator and the weakened variants of
the AES-128 and Magma block ciphers.
The runtime estimations of the constructed attacks push the state of the art.

This paper is organized as follows.
First, the next section introduces the notation used throughout the paper and
as well briefly describes the problem being studied.
Second, a new class of Inverse Backdoor Sets is introduced followed by a
discussion on how to use them when constructing guess-and-determine attacks on
cryptographic functions.
Third, we propose the resistance function used to compute runtime estimations
of the constructed guess-and-determine attacks as well as an automatic way to
minimize the resistance function.
%
%
Finally, the paper gives an overview of the preliminary experimental results of
applying the proposed technology to cryptanalysis of the Trivium, AES, and
Magma ciphers.

\section{Preliminaries}\label{sec:prelim}

Definitions standard in propositional satisfiability (SAT) are
assumed~\cite{biere-handbook09}.
In what follows, $C$ denotes an arbitrary propositional formula in
\emph{conjunctive normal form} (CNF), i.e.\ it is a conjunction of clauses.
A clause is a disjunction of literals while a literal is either a Boolean
variable $x$ or its negation $\neg{x}$.
Whenever convenient, CNF $C$ is defined over a set of variables $V$.

An assignment is a mapping $\alpha: V\rightarrow \{0,1\}$, which satisfies
(unsatisfies, resp.) a Boolean variable $x\in V$ if $\alpha(x)=1$
($\alpha(x)=0$, resp.).
Assignments defined for all variables of $V$ are called \emph{complete} while
assignments defined for a subset $B\subset V$ are called \emph{partial}.
Assignments can be extended in a natural way for literals ($l$) and clauses
($c$):
\begin{equation*} \renewcommand{\arraycolsep}{0.4em} \begin{array}{cc}
\alpha(l)=\left\{ \renewcommand{\arraycolsep}{0.1em} \begin{array}{ll}
\alpha(x), & \text{ if } l=x \\ 1-\alpha(x), & \text{ if } l=\neg{x} \\
\end{array} \right.  & \alpha(c)=\text{max}\{\alpha(l)\,|\,l\in c\} \end{array}
\end{equation*}
If assignment $\alpha$ satisfies every clause $c\in C$ then formula $C$ is said
to be \emph{satisfiable} and $\alpha$ is called a \emph{model} for $C$.

Whenever convenient, an assignment $\alpha$ to variables $V$ is denoted by the
corresponding set of variable values, i.e.\ $\alpha\in\{0,1\}^{|V|}$.
Hereinafter and following~\cite{williams-ijcai03}, assignments are used to
\emph{substitute} the corresponding variables with their values followed by the
formula \emph{simplification}.
In particular, given a subset of variables $B\subseteq V$ and an arbitrary
partial assignment $\beta\in\{0,1\}^{|B|}$, the result of substituting the
variables of $B$ with their corresponding values of $\beta$ in formula $C$ is
denoted by $C[\beta/B]$.
In a similar way, given two subsets of variables $B\subseteq V$ and
$B'\subseteq V$, $B\cap B'=\emptyset$, and assignments $\beta$ and $\beta'$, a
simplified CNF formula with variables $B$ and $B'$ assigned by $\beta$ and
$\beta'$ is denoted by $C[\beta/B,\beta'/B']$.

A large spectrum of various combinatorial problems can be reduced to SAT
effectively.
A number of cryptanalysis problems belong to this spectrum.
This paper considers the cryptanalysis problems in the context of the general
inversion problem for \emph{discrete functions}.
Hereinafter, a discrete function is meant to be a total function
$F:\{0,1\}^{*}\rightarrow\{0,1\}^{*}$ specified by some algorithm $M$.
We consider only discrete functions specified by polynomial-time algorithms.
An arbitrary polynomial-time algorithm defines the following family of
functions
\begin{equation}
  \label{eq1}
  f:\{0,1\}^{n}\rightarrow\{0,1\}^{m}, n\in \mathbb{N}_1.
\end{equation}

\begin{definition}[Inversion Problem] \label{def:inv}
  Given a number $n\in\mathbb{N}_1$ and some $\gamma\in Range\,f$, the
  \emph{inversion problem} for function $f$ (see \eqref{eq1}) defined by a
  program $M$ consists in computing $\alpha\in\{0,1\}^n$ s.t.\
  $f(\alpha)=\gamma$.
\end{definition}

Observe that many cryptanalysis problems can be considered in this context.
Suppose that given a \emph{secret key} $\alpha\in\{0,1\}^n$, $f$ generates a
pseudorandom sequence (generally speaking, of an arbitrary length), which is
later used to encrypt a \emph{plaintext} with bitwise XOR.
Such sequence is called a \textit{keystream}.
A fragment of the plaintext and the corresponding \emph{ciphertext}
identifies a fragment of the keystream, i.e.\ some word $\gamma\in Range f$,
for which we can consider the inversion problem: to find $\alpha\in\{0,1\}^n$
s.t.\ $f(\alpha)=\gamma$.
With respect to \emph{cryptographic keystream generators}, the problem
corresponds to the so-called \emph{known plaintext attack} --- the
cryptanalysis of the generator based on the known
plaintext~\cite{menezes-handbook96}.

Inversion of \eqref{eq1} can be reduced to SAT by considering a Boolean circuit
$S_f$ that defines function $f$.
Each input of $S_f$ is associated with a distinct Boolean variable $x_i\in X$,
$i\in [n]$, i.e.\ $|X|=n$.
We say that $X$ encodes the input of function $f$.
Analogously, the output of function $f$ is denoted by distinct Boolean
variables $Y=\{y_1,\ldots,y_m\}$, $X\cap Y=\emptyset$.
Circuit $S_f$ can be encoded into a CNF formula $C_f$ using auxiliary
variables~\cite{tseitin-scmml70}

Let $C_f$ be a CNF formula constructed for a circuit $S_f$ as described above
and let $\gamma$  be a partial assignment for variables $Y$ s.t.\ $\gamma\in
Range\,f$.
It can be observed that $C_f[\gamma/Y]$ is satisfiable and from any of its
models one can effectively extract a partial assignment $\alpha\in\{0,1\}^n$
s.t.\ $f(\alpha)=\gamma$.

Given all described above, the basic idea of SAT-based cryptanalysis is as
follows.
First, given a known algorithm $M$, a fixed $n\in\mathbb{N}_1$ and a fixed
$\gamma\in Range\,f$, construct a CNF $C_f[\gamma/Y]$.
Second, apply a SAT solving algorithm to $C_f[\gamma/Y]$.
Third, assuming that it manages to decide the formula within a reasonable time,
extract $\alpha\in\{0,1\}^n$ from the computed model of $C_f[\gamma/Y]$.

Note that although encoding cryptographic algorithms into SAT seems
conceptually simple, in practice it can be time consuming and non-trivial.
For our purposes it is important to know the input and output variables in a
constructed SAT instance. Also, the encodings must be arc consistent. Therefore,
for this purpose we use an automatic translator
Transalg~\cite{DBLP:conf/ecai/OtpuschennikovS16}, which satisfies all the requirements mentioned above.
%
To solve the constructed SAT instances, we employ the state-of-the-art
 \emph{conflict-driven clause learning} (CDCL) SAT solvers~\cite{jpms-cdcl09}.

Let us now proceed to the notion of \emph{Backdoor Set} for SAT.
The concept of Backdoor Sets and \emph{Strong Backdoor Sets} in the context of
CSP and SAT was first strictly formalized in~\cite{williams-ijcai03}.

\begin{definition}[Strong Backdoor Set] \label{def:strong}
  Let $C$ be an arbitrary CNF formula over a set of variables $V$, and let $A$
  be a polynomial-time algorithm.
  A non-empty set $B$, $B\subseteq V$, is a \emph{Strong Backdoor Set} for $C$
  w.r.t.\ algorithm $A$ if for each $\beta \in \{0,1\}^{|B|}$ algorithm $A$
  decides formula $C[\beta/B]$.
\end{definition}

A simple, but important class of Strong Backdoor Sets is formed by Strong Unit Propagation Backdoor Sets (SUPBS) \cite{williams-ijcai03}. In this case algorithm $A$ is Unit Propagation \cite{dowling-jlogp84}.  For example the set of input variables  $X$ in $C_{f}$ is a SUPBS.

\section{Backdoor Sets for Non-polynomial-time and/or Incomplete Algorithms}
\label{sec:inv}

The intuition behind Definition~\ref{def:strong} is that computing a Strong Backdoor
Set for a SAT problem enables one to \emph{predict} the runtime of the complete
solving process exploiting the backdoors.
Indeed, it can be computed as the number of all possible partial assignments
for the backdoor variables multiplied by the time complexity of algorithm $A$.

Unfortunately, Strong Backdoors are typically either extremely hard to find, or
do not pose any interest because of their size (and, thus, the number of
subproblems to solve).
Here the main restriction is, in fact, the requirement that algorithm $A$ must
have a polynomial complexity.
Instead of $A$, one may want to use some complete algorithm $A'$ that has an
exponential worst-case scenario time complexity but works well on average.
As an example, one could use an NP-oracle, e.g. SAT- or CSP-solving algorithm,
as $A'$.
Clearly, there are CNF formulas, which can be hard to decide by
state-of-the-art SAT solvers used as blackboxes, but can still be solved by the
partitioning approach, as described in~\cite{hyvarinen-phd11}.

%
\begin{definition}[Non-deterministic Oracle Backdoor Set] \label{def::NDOBS}
  Let $A$ be a complete algorithm and let $C$ be an arbitrary CNF formula over
  a set of variables $V$.
  A non-empty set $B$, $B\subseteq V$, is a \emph{Non-deterministic Oracle
  Backdoor Set (NOBS)} w.r.t. algorithm $A$ if the total running time of $A$ given
  formulas $C[\beta/B]$, $\beta \in \{0,1\}^{|B|}$, is less than the running
  time of $A$ on the original formula $C$.
\end{definition}

NOBS can be seen as a special case of plain partitioning \cite{hyvarinen-phd11}.
It is such a set of variables that enables
solving a problem faster (by decomposing or partitioning it via the backdoor
variables) than if the same algorithm is applied to the original formula.
The "non-deterministic oracle" part refers to the fact that we can use any
algorithm for solving problems from NP.
The Cube-and-Conquer solvers~\cite{heule-hvc11} implicitly use an NOBS-like idea: they employ CDCL solvers to explore
branches of the search tree chosen by e.g.\ a lookahead solver. In some cases it
leads to spectacular results~\cite{heule-sat16}.

Note that by using NOBS instead of Strong Backdoors we lose the ability to analytically
evaluate the time required for solving the problem.
In practice, this disadvantage can be alleviated as follows: one can solve a
reasonably small portion of subproblems and use their average solving time to
extrapolate how long it will take to solve \emph{all} subproblems.

Note that the Backdoor Sets of the introduced type, as well as Strong Backdoor sets
from \cite{williams-ijcai03} are essentially \emph{constraint-agnostic} and can
be adapted to almost any variants of Constraint Satisfaction Problems and
solving algorithms.

At this point, it is also possible to discard the requirement that the algorithm is complete.
In the reminder of the paper we introduce a new type of backdoor sets following this paradigm in the context of guess-and-determine attacks on cryptographic functions and analyze how it can be applied in practice.

\subsection{Guess-And-Determine attacks on Cryptographic Functions}
This section describes the interconnection between the proposed Backdoor Sets
and guess-and-determine attacks on cryptographic functions.
Let us first discuss the properties of a cryptographic function.
Typically, given a function $f(\cdot)$ and an arbitrary $\alpha$ as an input,
the algorithm that computes $\gamma=f(\alpha)$ has polynomial time complexity.
It is justified by the fact that nowadays the amount of data to be encrypted is
increasing every year and, thus, the performance of an encrypting algorithm is
vital.
The second characteristic of a cryptographic function is that its inversion,
i.e.\ computing $\alpha$ s.t.\ $f(\alpha)=\gamma$ given $\gamma$, must be
extremely hard (ideally, infeasible).
Here we are mostly interested in constructing runtime estimations when solving
inversion problems for cryptographic functions by exploiting characteristics of
different Backdoor Sets.
Let us consider the relation between the known and the proposed Backdoor
Sets, as well as guess-and-determine attacks.

The basic idea of a guess-and-determine attack is quite simple.
Assume that we have a formula $C$ describing a cryptanalysis problem and $V$ is
a set of variables of this formula.
Then a guess-and-determine attack is as follows: (i) ``guess'' values of the
variables from a subset $B$, $B\subseteq V$, and then (ii) ``determine'' using
a relatively fast algorithm whether or not the guess was correct.
Trivially, this approach implies doing exhaustive search over $2^{|B|}$
possible assignments to variables of $B$.
Therefore, in light of cryptanalysis any Backdoor Set mentioned above
corresponds to a guess-and-determine attack.
For example, one can observe that a set of variables encoding an input of a
cryptographic function can serve as its trivial Strong Backdoor.
The corresponding guess-and-determine attack is a \emph{brute force} attack on
the cryptographic function.
Also, for specific ciphers there exist less trivial Strong Backdoors.
For instance, \cite{anderson-nc94} proposed such a Strong Backdoor for the A5/1 keystream generator. Once the variables from the corresponding set are assigned, SAT for the resulting subformula can be solved by Unit Propagation.

Non-deterministic Oracle Backdoor Sets correspond to cryptographic attacks, where NP-oracles are used to solve subproblems constructed for a specific Backdoor set.
For instance, very similar approach was applied to
cryptanalysis of the widely known Bivium and Trivium ciphers~\cite{eibach-sat08,Semenov2016,eibach-mcs10,huang-africacrypt11}.

The main goal of the present paper is to develop a method for automatic
construction of guess-and-determine attacks on cryptographic functions, that
would rely on exploiting backdoor sets in a CNF formula encoding a
cryptanalysis problem.
Unfortunately, applying Non-deterministic Oracle
Backdoor sets to constructing formally justified cryptographic attacks has a
few major drawbacks. The main one is that, by definition, for each new output
of the considered function it is necessary to construct a new Backdoor set.
Ideally, we need such a backdoor set that would retain its properties through
all possible pairs of input-output of the considered cryptographic function.

Taking into account all of the above, let us introduce a new type of Backdoor
sets, designed specifically for solving inversion problems for cryptographic
functions.
Consider an arbitrary function $f$ of form \eqref{eq1}.
First, encode $f$ to a CNF formula $C_f$ s.t.\ sets of variables $X$ and $Y$
represent input and output of $f$.
Let $B$ be an arbitrary subset of $X$ and consider some
$\alpha\in\{0,1\}^{|X|}$.
By $\beta(\alpha)$ we denote a partial assignment to variables of $B$ extracted
from a specific $\alpha$.
We say that $\beta(\alpha)$ is induced by assignment $\alpha$.

Define a uniform distribution over $\{0,1\}^n$.
With each $\alpha$ randomly selected from $\{0,1\}^n$ we associate assignments
$\beta(\alpha)$ (induced by $\alpha$) -- values of the Backdoor variables, and
$f(\alpha)$ -- an assignment of output variables from $Y$.
Next, consider a simplified CNF formula $C_f[f(\alpha)/Y,\beta(\alpha)/B]$.
Clearly, this formula is satisfiable, and from each of its models one can
extract the corresponding partial assignment $\alpha\in\{0,1\}^n$.
Let $A$ be an arbitrary SAT solving algorithm.
For each triple $(\alpha,f(\alpha),\beta(\alpha))$, the running time of $A$ on
formula $C_f[f(\alpha)/Y,\beta(\alpha)/B]$ is denoted by $T_A\left(C_f,
B,\alpha\right)$.
Let $t$ be a parameter that takes only positive values. For a fixed $t$
consider the following value:
\begin{equation} \label{eq2}
  P_B(t)=\frac{\#\left\{\alpha\in\{0,1\}^n:T_A\left(C_f, B,\alpha\right)\leq
  t\right\}} {2^n}
\end{equation}

The numerator of \eqref{eq2} depicts a number of such $\alpha\in\{0,1\}^n$ for
which CNF formula $C_f[f(\alpha/Y,\beta(\alpha)/B]$ is decided by $A$ in time
$\leq t$.
The denominator of \eqref{eq2} is the number of all $\alpha$.
Therefore, \eqref{eq2} is essentially a probability of the following event: a
randomly selected $\alpha\in\{0,1\}^n$ induces such assignments $\beta(\alpha)$
and $f(\alpha)$ that $A$ decides $C_f[f(\alpha/Y,\beta(\alpha)/B]$ in time
$\leq t$.

\begin{definition}[Inverse Backdoor Set] \label{def:IBS}
  An arbitrary non-empty set $B$, $B\subseteq X$, $|B|=s$, with properties
  described above is called an \emph{Inverse Backdoor Set (IBS)} with
  parameters $(s,t,P_B(t))$ for $C_f$ w.r.t. algorithm $A$.
\end{definition}
Note that probability \eqref{eq2} is not tied to a specific output of $f$.
Therefore, any IBS can be used to mount a general guess-and-determine attack on a
cryptographic function.

Now let us describe a guess-and-determine (in the following \emph{G-a-D})
attack strategy that corresponds to IBS $B$ with parameters $(s,t,P_B(t))$.

\begin{definition}[elementary G-a-D attack based on IBS $B$]
  \label{def:IBS-attack}
  Consider the inversion problem for function $f$ of form \eqref{eq1} given an
  arbitrary $\gamma\in Range\,f$.
  \begin{enumerate}
    \item Assume that $\gamma = f(\alpha)$ for some $\alpha\in \{0,1\}^n$.
    \item Let $B$ be an IBS with parameters $(s,t,P_B(t))$ and
      $\beta\in\{0,1\}^s$ is an assignment to variables of $B$.
    \item Construct CNF $C_f[\gamma/Y,\beta/B]$  and run a SAT solver $A$ on
      it.
    \item If the runtime of $A$ on this SAT instance exceeds $t$, interrupt the
      solving process and move to another $\beta$.
    \item For $\beta=\beta(\alpha)$, an algorithm $A$ will find a model for
      $C_f[\gamma/Y,\beta/B]$ in time $\leq t$ with probability $P_B(t)$.
      This means that $A$ will compute $\alpha$ s.t.\ $f(\alpha)=\gamma$.
  \end{enumerate} \end{definition}

It is possible that the analysis of a $\gamma$ lead to \emph{no result} (since on each
formula $C_f[\gamma/Y,\beta/B]$ the runtime of $A$ exceeded $t$). In that case due to
the cryptographic context, we can consider another point from $Range\,f$ that is
different from $\gamma$. For example, in application to block ciphers it means that we need
another pair input-output, where input is encrypted using the secret key applied to produce $\gamma$.
%
Thus, we have the following iterative guess-and-determine attack.

\begin{definition}[G-a-D attack, based on IBS $B$].
\label{def:GDA-IBS-attack}
  Consi\-der the inversion problem for a function $f$ of form \eqref{eq1}.
  \begin{enumerate}
    \item Let $\gamma^1,\ldots,\gamma^r$ be observed outputs of function $f$.
      These outputs correspond to inputs $\alpha^1,\ldots,\alpha^r$.
    \item Let $B$ be some IBS with parameters $(s,t,P_B (t))$, $P_B(t)>0$.
    \item A guess-and-determine attack based on IBS $B$ consists in successive
      application of elementary attack, as described in Definition~\ref{def:IBS-attack}, to outputs $\gamma^1,\ldots,\gamma^r$.
    \item The attack is said to be successful if for at least one
      $j\in\{1,\ldots,r\}$ the corresponding inversion problem for $f$ is
      solved.
  \end{enumerate}
\end{definition}

It is easy to observe that the probability of success of a proposed
guess-and-determine attack on the set of outputs $\gamma^1,\ldots,\gamma^r$ is
\begin{equation} \label{eq3}
  P_r^{*}=1-\left(1-P_B(t)\right)^r
\end{equation}
Thus, assuming that $P_B(t)>0$, the probability of success of this attack
converges to $1$ with the increase of $r$.

In contrast to guess-and-determine attacks based on Strong Backdoor sets, 
it is unclear at first glance how to
estimate the running time of an IBS-based attack.
However, for each $B$ its size $s$ is known, and a value of $t$ is specified
beforehand.
Therefore, the main question regarding IBS-based attacks is how to calculate
the value of probability $P_B(t)$ within a reasonable time.
The next section will describe the technique for estimating $P_B(t)$ for fixed
backdoors $B$ and time $t$.
It will also introduce the concept of resistance function for $B$, whose values
are estimations of complexity of the corresponding guess-and-determine attacks.
Finally, it will reduce the problem of computing an IBS corresponding to
attacks with low complexity to minimization of the resistance function over a
Boolean hypercube.

\section{Dealing with IBS}
First, let us discuss how to compute the runtime estimation for a specific IBS.

Assume that when inverting function \eqref{eq1} we consider an arbitrary IBS $B$ with parameters $(s,t,P_B (t))$. First, we need to compute the value of probability $P_B (t)$. It can not be done effectively, because according to \eqref{eq2} for this purpose we would need to know the number of such
$\alpha\in\{0,1\}^n$, for which the runtime of $A$ on CNFs $C_f [f(\alpha)/Y,\beta(\alpha)/B]$ does not exceed $t$. However, we can  use the Monte-Carlo method \cite{metropolis-jasa49} to estimate $P_B(t)$.
The general idea of the Monte Carlo method is simple. Let $\xi$ be a random variable. Assume that its expected value  $E[\xi]$ and variance $Var(\xi)$ are both finite, but unknown and can not be computed in reasonable time. Nevertheless, we want to know ``what to expect'' from $\xi$, i.e. to estimate its expected value $E[\xi]$. For this purpose, a \textit{random sampling} is first performed by making $N$ independent observations of $\xi$: $\xi^1,\ldots,\xi^N$. Then from the \emph{Central Limit theorem} \cite{feller-pth71} it follows that if the size of the random sample $N$ is large enough, then the sample mean shown in \eqref{eq5}
\begin{equation}
\label{eq5}
\overline{\xi}=\frac{1}{N}\sum_{j=1}^N\xi^j
\end{equation}
can be considered as a good approximation of $E[\xi]$.

Let us illustrate this concept. Assume that we have a CNF $C$, a Non-deterministic Oracle Backdoor Set $B$ and use SAT solver $A$ to solve simplified subproblems $C[\beta/B]$, $\beta\in\{0,1\}^{|B|}$. Assume that $\beta$ is chosen from $\{0,1\}^{|B|}$ according to uniform distribution. Then the runtime of $A$ given an arbitrary subproblem $C[\beta/B]$ can be treated as an observation of a random variable $\xi$. Here it is assumed that the goal is to construct a runtime estimation for solving $C$ by exploiting $B$, since the solving itself would require too much resources. For this, we randomly generate $N$ assignments $\beta^1,\ldots,\beta^N$ of variables from $B$ and track the runtime of $A$ on $C[\beta^j/B]$, $j\in\{1,\ldots,N\}$: $\xi^j=T_A(C[\beta^j/B])$. The sample mean $\overline{\xi}$ computed according to \eqref{eq5} can be considered as an estimation of an average time required by $A$ to solve an
\emph{average} subproblem $C[\beta/B]$. By multiplying it by $2^{|B|}$, an estimation of the total time to solve all $C[\beta/B]$, $\beta\in\{0,1\}^{|B|}$ is constructed.

Now let us return to the problem of estimating $P_B(t)$. Assume that $B$ is an IBS. The goal given randomly chosen input-output pairs $\gamma=f(\alpha)$ is to estimate what portion of inversion problems $C_f[\gamma/Y,\beta(\alpha)/B]$ can be solved by algorithm $A$ within time limit $t$. For this purpose, define a random variable $\xi$ as follows. For a randomly selected assignment $\alpha\in\{0,1\}^n$ the value of $\xi=\xi_A(B,\alpha,t)$ is equal to $1$ if $A$ solves SAT for CNF $C_f[f(\alpha)/Y,\beta(\alpha)/B]$ in time $\leq t$ and $0$ otherwise. From \eqref{eq2} it follows that $\xi$ takes values of $1$ and $0$ with probabilities $P_B (t)$ and $1-P_B(t)$, respectively. Thus, $E[\xi]=P_B (t)$. Since $E[\xi]$ and $Var(\xi)$ are both finite, and thus formal requirements of the Monte Carlo method are satisfied, we can use the following scheme to estimate $P_B(t)$ for a fixed $t$.
\begin{enumerate}
\item Generate a random sample of assignments $\alpha^1,\ldots\alpha^N$, $\alpha^j\in\{0,1\}^n$, $j\in\{1,\ldots,N\}$.
\item For each $\alpha^j$, construct $f(\alpha^j)$ and $\beta(\alpha^j)$, construct CNF $C_f[f(\alpha^j)/Y,\beta(\alpha^j)/B]$ and run $A$ on it.
\item If within time limit $t$, algorithm $A$ computes a model, then $\xi^j=1$, otherwise $\xi^j=0$.
\item Compute sample mean $\overline{\xi}$ according to \eqref{eq5}.
\end{enumerate}
Obviously, the larger the size of a random sample, the better we can construct the approximation of $P_B(t)$. Now that it is clear how to estimate $P_B(t)$ it is possible to move to the problem of estimating the runtime of an IBS-based attack.

\subsection{Estimating Runtime of IBS-based attacks}

Let us get back to the notion of guess-and-determine attack defined in Definition~\ref{def:GDA-IBS-attack}.
Assume that following Definition~\ref{def:GDA-IBS-attack}, we successively apply an elementary guess-and-determine attack based on $B$ to analyze several outputs $\gamma^1,\ldots,\gamma^r$. Therefore, the total runtime of a corresponding attack is $r\times t\times 2^{|B|}$. However, it is necessary to estimate $r$: how many outputs should be analyzed for a considered guess-and-determine attack to yield a result. For this purpose, we use formula \eqref{eq2} under the assumption that the probability of success equal to $95\%$ meets our needs. So it is necessary to choose $r$ in such a way that $P_r^*\geq 0.95$. Since all considered probabilistic spaces are finite and without loss of generality, assume that $P_B (t)$ is a positive rational number $1/q$, $q\geq 1$. Taking into account the fact that for large $q$ it holds that $\left(1-\frac{1}{q}\right)^q\approx e^{-1}\approx 0.3678$, it follows from \eqref{eq2} that the probability $P_r^*$ exceeds $0.95$ when $r\approx \frac{3}{P_B (t)}$.  Using sample mean $\overline{\xi}$ as an estimation of $P_B(t)$ we derive  the following definition.

\begin{definition}
Assume that there is an IBS $B$ with parameters $(s,t,P_B (t))$. The following function
\begin{equation}
\label{eq6}
G(B)=2^s \times t \times \frac{3}{\overline{\xi}},
\end{equation}
is referred to as \emph{resistance function}.
\end{definition}
Here, ``resistance'' should be understood as \textit{cryptographic resistance to IBS-based attacks}.
By definition, the value of a resistance function for an IBS $B$ is the runtime estimation of the corresponding guess-and-determine attack based on $B$. In the following subsection we propose the algorithm that can be used to find a good IBS via minimization of the resistance function over all possible IBSes. We use this algorithm to automatically construct IBS-based attacks on several state-of-the-art cryptographic functions.

\subsection{Minimizing the Resistance Function}
The minimization problem for the resistance function is complicated by the fact that this function is not specified analytically. Another distinctive feature of this function is that computing its value in some point requires tracking the runtime of the employed algorithm $A$. In fact, tackling such functions is a common problem in the area of blackbox optimization, where a function is treated as a blackbox, which given some input produces some output according to its (unknown) design.

Note that computing the resistance function in one point is actually quite expensive: if we use a random sample of size $N$ and time limit $t$, in the worst case it takes $N\times t$ to compute its value. Therefore, first, it is undesirable to compute the function in a point of the search space more than once. Second, in practice, the number of points to be processed is limited by the amount of available computational resources. Thus, it is convenient to minimize the resistance function by a combination of \emph{Tabu search} \cite{glover-ts97} heuristics with some variant of local search. Tabu search uses a \emph{Tabu list}, where the algorithm stores all points for which the function value was computed. Therefore, before computing the value of the function in the next point, it first checks whether or not this point is already in the Tabu list.

Let us formally describe the minimization algorithm. Assume that there is a CNF formula $C_f$ specifying some inversion problem. Let $X=\{x_1,\ldots,x_n\}$ be a set of Boolean variables encoding the input of $f$. An arbitrary subset $B$ of $X$ can be described using Boolean vector $\chi=\chi(B)=(\chi_1,\ldots,\chi_n)$, where for $i\in\{1,\ldots,n\}$:
$$
\chi_i=\left\{
\begin{array}{l}
1,x_i\in B\\
0,x_i\notin B
\end{array}
\right.
$$
Thus, the search space can be defined as an $n$-dimensional Boolean hypercube $E^n=\{0,1\}^n$. For an arbitrary point $\chi\in E^n$, a neighborhood $Nh(\chi)$ of radius $R$ is defined as a set of such points $\chi'$, $\chi^{'} \in E^n$, that $d_H(\chi,\chi')\leq R$, where $d_H(\chi,\chi')$ stands for Hamming distance between $\chi$ and $\chi'$ 
(in this work we considered radius $R=1$).

Note that set $B=X$ is an IBS with $P_B (t)=1$ for some small $t$. Indeed, in this case for each $\alpha\in \{0,1\}^n$ it follows that $\beta(\alpha)=\alpha$, and, therefore, SAT for an arbitrary CNF formula $C_f [f(\alpha)/Y,\alpha/X]$ can be solved using only Unit Propagation. This means that it is always possible to start the process of minimizing \eqref{eq6} from the point $\chi_{start}=1^n$  (i.e. the vector of all $1$s), corresponding to $X$.

Algorithm~\ref{alg:tabu} shows the pseudo-code of the algorithm.
We denote the current point, a neighborhood of which is being processed, as $\chi_{center}$, the current Best Known Value of \eqref{eq6} as $G_{best}$, the corresponding $\chi$ as $\chi_{best}$.
Function \texttt{Resistance($\chi$,t)} computes the value of the resistance function for Backdoor $B$, corresponding to point $\chi$, with the time limit $t$. Function \texttt{UpdateTabuList($\chi$)} adds the corresponding point to the Tabu list. The algorithm uses function \texttt{GetNewCenter()} to exit local minima. Function \texttt{TimeExceeded()} checks if the time limit of the algorithm is exceeded (in our experiments we used the time limit of 1 day).

\begin{algorithm}
 \DontPrintSemicolon
 \SetKwData{false}{false}
 \SetKwData{true}{true}
 \SetKwData{NewOptimum}{NewOptimum}
 \SetKwData{TabuList}{TabuList}
 \SetKwFunction{UpdateTabuList}{UpdateTabuList}
 \SetKwFunction{GetNewCenter}{GetNewCenter}
 \SetKwFunction{TimeExceeded}{TimeExceeded}
 \SetKwFunction{ConstructIBS}{ConstructIBS}
 \SetKwFunction{UncheckedPoint}{UncheckedPoint}
 \SetKwFunction{Resistance}{Resistance}
 \caption{Tabu search algorithm for minimization of the resistance function}
 \label{alg:tabu}
	\KwIn{CNF formula $C_f$, time limit $t$}
	\KwOut{$\chi_{best}$ with the runtime estimation $G_{best}$}
    $\chi_{best} \gets \chi_{center} \gets 1^n$\;
    $G_{best} \gets G(1^n)$\;
	\Repeat{\TimeExceeded{}} {
		\NewOptimum $\gets$ \false\;
        \Repeat{All points in $Nh(\chi_{center})$ are checked}{
			$\chi \gets$ \UncheckedPoint($Nh(\chi_{center}))$\;
			$g \gets \Resistance(\chi, t)$\;
			\UpdateTabuList{$\chi$}\;
			\If{$g < G_{best}$}{
				$\langle \chi_{best}, G_{best} \rangle \gets \langle \chi, g \rangle$\;
				\NewOptimum $\gets$ \true\;
			}
		}
		\lIf{\NewOptimum} {
			$\chi_{center} \gets \chi_{best}$
		}
		\lElse{
			$\chi_{center} \gets \GetNewCenter()$
		}
	}
\Return{$\langle \chi_{best}, G_{best} \rangle$}\;
\end{algorithm}

\section{Preliminary Experimental Results}\label{sec::exp}
All experiments were run on 10 nodes of a computing cluster, each node being
equipped with two Intel~Xeon~E5-2695~v4 CPUs and 128 GB RAM.
All presented runtime estimations are scaled to one core of the aforementioned
CPU.
During the resistance function minimization, we used random samples of size $1000$.
This choice represents a compromise: on the one hand, the larger the sample size is, the more
accurate the results of the Monte Carlo method are. On the other hand, if we use samples of
a larger size, the performance of traversing the search space is greatly hampered.
However, in order to ensure the consistency of the obtained results, for every computed
IBS $B_{best}$, the value of the resistance function was recomputed using random samples
of increasing size up to $100000$.

When searching for best IBSs, the values of $P_B$ and $t$ typically were in intervals
$P_B\in [0.05,1]$, $t\in [1s,200s]$. The intuition here is simple: if $P_B$ is too small,
then the impact of randomness on the resistance function's value is too large. During
the empirical evaluation, $P_B<0.05$ turned out to lead to significant deviations of the resistance
function's values computed using random samples of increasing size. As for $t$, when it is
small then SAT solvers cannot fully employ all their capabilities; and when $t$ is too large,
the proposed method becomes computationally too expensive.

Experimental results were obtained using the ROKK SAT solver~\cite{yasumoto-rokk}.
This non-standard choice is a result of a rigorous evaluation
comparing top CDCL solvers from SAT competitions 2014--2016 (on the formulas studied in this work), including
ROKK, lingeling, Cryptominisat, etc.  Surprisingly, ROKK turned out to be the
winner.

Below we show the results of constructing guess-and-determine attacks on 3
widely known symmetric ciphers.
The first one is the Trivium stream cipher~\cite{canniere-isc06}.
It is one of the winners of eSTREAM, which is a project aimed at identifying
new stream ciphers suitable for widespread use.
We considered the problem of recovering initial states of the Trivium registers
(288 bits in total) for a known 300-bit keystream fragment.

The second benchmark is the problem of finding a secret key given three
blocks of known plaintext (3KP) for the reduced version of the AES-128 block cipher.
3KP are needed to ensure the probability \eqref{eq3} to be $\geq 95\%$.
AES-128 is originally a 10-round \emph{Substitution Permutation
Network}.
The full-round AES is one of the most cryptographically resistant
state-of-the-art ciphers.
The majority of the known attacks with the runtime significantly smaller than
that of the exhaustive search are designed only for AES with reduced numbers of
rounds.
We studied the 2.5-round version of AES-128 (following the notation
of~\cite{bouillaguet-crypto11}, ``x.5r'' means $x$ full rounds and the final
round).

Finally, we studied the problem of finding a secret key given 12 blocks of the known plaintext (\emph{12KP}) and the corresponding ciphertext for the reduced version of the Magma cipher (GOST~28147-89).
This cipher was used in the USSR and Russia from 1989 to 2015.
Originally, Magma is a 32-round cipher based on the \emph{Feistel network}
architecture.
Similar to AES, significant improvements in cryptanalysis of Magma compared to
the brute force attacks are known only for reduced-round variants of the Magma
cipher.
We studied the 8-round variant of Magma. For ensuring \eqref{eq3} to be
$\geq 95\%$, we need 12 blocks of known plaintext.

The estimations of the guess-and-determine attacks constructed by our method
for the considered ciphers are showed in Table~\ref{tab:res}. The sizes of the
corresponding IBS sets are $|B|=131$ (out of $|X|=288$) for Trivium, $|B|=63$ (out of $|X|=256$) for Magma and $|B|=42$ (out of $|X|=128$) for AES.

\paragraph{Trivium.} In~\cite{borghoff-sac10} a guess-and-determine attack on
Trivium is a result of solving a discrete optimization problem over a Boolean
hypercube.
In contrast to our generic method, the approach of Borghoff et\,al.\
targets only the cryptanalysis equations for Trivium and does not apply a
general technique or method for constructing guess-and-determine attacks on a
large class of cryptographic functions.
Also, in~\cite{borghoff-sac10} the effectiveness of a guess-and-determine
attack is estimated in a completely different way if compared with the proposed
resistance function. Borghoff et\,al.\ proposed an attack with the estimation of $4.31{e}{+55}$ seconds.
A guess-and-determine attack on Trivium with the smallest runtime estimation is
described in~\cite{huang-africacrypt11}, which uses the ``Characteristic Set
method'' (CS-method) for solve the Trivium cryptanalysis equations.
Although we acknowledge that the method proposed in~\cite{huang-africacrypt11}
outperforming the approach of the present paper (applied to Trivium) is
somewhat discouraging, note that~\cite{huang-africacrypt11} did not propose 
a versatile automatic procedure applicable to other ciphers (in contrast to our approach of resistance function minimization).
%

\begin{table}[t]
\caption{Estimated hardness of the guess-and-determine attacks for weakened
variants of Trivium, AES, and Magma compared to Previous Best Attacks (PBA),
i.e.\ see~\cite{huang-africacrypt11} for Trivium, \cite{bouillaguet-crypto11} for
AES, and \cite{courtois-tatra12} for Magma.
Time complexity is measured in seconds scaled to one core of the 
Intel~Xeon~E5-2695~v4 CPU while memory is measured in bits. Here by ``negligible''
we mean the amount of memory, which is standard for a modern PC.}

\begin{center}
\begin{tabular}{p{1.4cm}p{1.5cm}p{1.5cm}p{2cm}}
\toprule
Cipher & Time & Memory & Reference
\\
\midrule
Trivium & $2.04\mathrm{e}{+41}$ & negligible & present paper\\
\cmidrule{2-4}
 & $2.50\mathrm{e}{+34}$ & negligible & PBA \\
\midrule
AES-128
 & $1.45\mathrm{e}{+15}$ & negligible& present paper\\
\cmidrule{2-4}
 & $3.08\mathrm{e}{+16}$ & $2^{80}$ & PBA \\
\midrule
Magma & $3.55\mathrm{e}{+22}$ & negligible & present paper\\
\cmidrule{2-4}
 & $1.17\mathrm{e}{+23}$ & negligible & PBA \\
\bottomrule
\end{tabular}
\end{center}
\label{tab:res}
\end{table}

\paragraph{AES-128.} To our best knowledge, the state-of-the-art
guess-and-determine attacks on AES-128 with a reduced number of rounds were
proposed in~\cite{bouillaguet-crypto11}.
Bouillaguet et\,al.\ considered a set of all possible sets of guessed bits as
a tree traversed in a way similar to the branch-and-bound method.
They claim that their method ``...~is
reminiscent of the DPLL procedure implemented in many SAT-solvers''.
However, their approach does not use SAT solvers.
Also, they do not estimate the runtime of a guess-and-determine attack similarly to what is done in the present paper: by analyzing the performance of an algorithm for solving a set of weakened cryptanalysis instances.
\cite{bouillaguet-crypto11} considered the  cryptanalysis of a truncated AES-128 (with 2.5 rounds), in which 2KP are analyzed.
The main disadvantage of this method consists in an enormous amount of memory consumption.
In all our estimations the amount of memory required is a tiny fraction of the required runtime.

\paragraph{Magma.} In \cite{courtois-tatra12} the Magma cipher was studied by SAT solvers.
To estimate the performance of guess-and-determine attacks, Courtois et\,al.\ introduced
the notions of SAT-immunity and
UNSAT-immunity, which, however, were not strictly formalized.
Our notion of resistance function can be seen as a \emph{concretization}
of the notion of SAT-immunity.
The attack of Courtois et\,al. (see its running time in
Table~\ref{tab:res}) was constructed as a result of a thorough analysis of the
Magma design features without using automatic algorithms for constructing
guess-and-determine attacks. Note, that in \cite{courtois-tatra12} 4KP were analyzed, while our attack requires 12KP in accordance with the aforementioned reasons. However, the runtime estimation of our attack is lower.

%

\section{Conclusions} \label{sec:conc}
The paper studies a new class of Backdoors Sets for SAT (Inverse Backdoor Sets,
IBS), which aims at facilitating efficient cryptographic attacks, namely
guess-and-determine attacks.
The values of the backdoor variables are used as bits to guess in the proposed
guess-and-determine attack.
The efficiency/hardness of the attack is defined as a value of a specific
resistance function, which is estimated statistically using the Monte-Carlo
method.
The idea of the proposed approach is to identify the best set of backdoor
variables subject to the hardness of the guess-and-determine attack, i.e.\ the
value of the resistance function, using a SAT solver.
Preliminary experimental results indicate that the proposed approach pushes the
state of the art in the estimating hardness of the guess-and-determine attack
for a number of weakened variants of the known encryption algorithms, namely AES and Magma.

The following lines of future work can be envisioned.
First, one can extend the proposed IBS-based guess-and-determine attacks taking
into account not only the variables encoding the algorithm's input but also the
auxiliary variables introduced by the encoding process.
Second, the new class of backdoors is general enough to be readily adapted to
other classes of hard SAT formulas, e.g.\ in the context of parallel and
distributed SAT solving, as well as other types of constraints.
Third, in order to estimate the hardness of a guess-and-determine attack based
on some backdoor set, it seems plausible to adapt known \emph{deterministic}
measures, e.g. to analyze ``current space complexity''~\cite{Ansotegui2008},
instead of running time.
%



\bibliographystyle{aaai}
\bibliography{refs_short}

\end{document}